\documentclass[runningheads]{llncs}
\usepackage{graphicx}
\usepackage{amsmath,amssymb} 
\usepackage{color}

\usepackage{multirow}
\usepackage{bm}
\usepackage{url}
\usepackage{array}
\usepackage{tabularx}
\usepackage{mathtools}
\usepackage{listings}

\usepackage{float}

\begin{document}

\newcommand{\point}{
    \raise0.7ex\hbox{.}
    }


\pagestyle{headings}

\mainmatter

\title{On the Choice of Tensor Estimation for Corner Detection, Optical Flow and Denoising} 

\titlerunning{Tensor Estimation for Corner Detection, Optical Flow and Denoising} 

\authorrunning{F. {\AA}str{\"o}m and M. Felsberg} 

\author{Freddie {\AA}str{\"o}m\inst{1,2} and Michael Felsberg\inst{1,2}} 

\institute{Computer Vision Laboratory, Link{\"o}ping University, Sweden \and  
 Center for Medical Image Science and Visualization (CMIV), Link{\"o}ping University
}

\maketitle

\renewcommand{\div}[1]{\operatorname{div} \left ( #1 \right )}
\newcommand{\tr}[1]{\mathrm{tr} \;  #1 }
\newcommand{\norm}[2]{|| #1 ||_{#2}}
\newcommand{\pd}[2]{\frac{\partial #1}{\partial #2}}
\newcommand{\eg}[0]{\emph{e.g. }}
\newcommand{\ie}[0]{\emph{i.e. }}
\newcommand{\dive}{\,\mathrm{div}}
\newcommand{\figwidth}{0.24}

\newcommand{\bxi}[0]{\bm{\xi}}
\newcommand{\bx}[0]{\bm{x}}

\newcommand{\paraheading}[1]{\vspace*{2mm}\noindent \textbf{#1}}

\newcolumntype{M}{>{\centering\arraybackslash}m{\dimexpr.25\linewidth-2\tabcolsep}}
\renewcommand{\labelitemi}{$\bullet$}

\begin{abstract}
	Many image processing methods such as corner detection, optical flow and iterative enhancement make use of image tensors. Generally, these tensors are estimated using the structure tensor. In this work we show that the gradient energy tensor can be used as an alternative to the structure tensor in several cases. We apply the gradient energy tensor to common image problem applications such as corner detection, optical flow and image enhancement. Our experimental results suggest that the gradient energy tensor enables real-time tensor-based image enhancement using the graphical processing unit (GPU) and we obtain 40\% increase of frame rate without loss of image quality.
\end{abstract}



\section{Introduction}
A drawback of many current state of the art image processing methods is the high computation requirements necessary to achieve high-quality results. As a consequence, the computational constraints limit the methods applicability as useful tools in real-time applications, implying that processing-pipelines operate on suboptimal image data. Specifically, the structure tensor introduced by F{\"o}rstner and G{\"u}lch~\cite{Forstner87} and Big{\"u}n and Granlund~\cite{diva2:274026} is an integral part of many image processing applications, such as corner detection~\cite{Harris88acombined,323794}, optical flow~\cite{Lucas:1981:IIR:1623264.1623280} and tensor-based image denoising~\cite{Weickert96anisotropicdiffusion}. 

\begin{figure*}[t]
   \centering
   \begin{minipage}{0.40\textwidth}
		\includegraphics[width=1.03\textwidth]{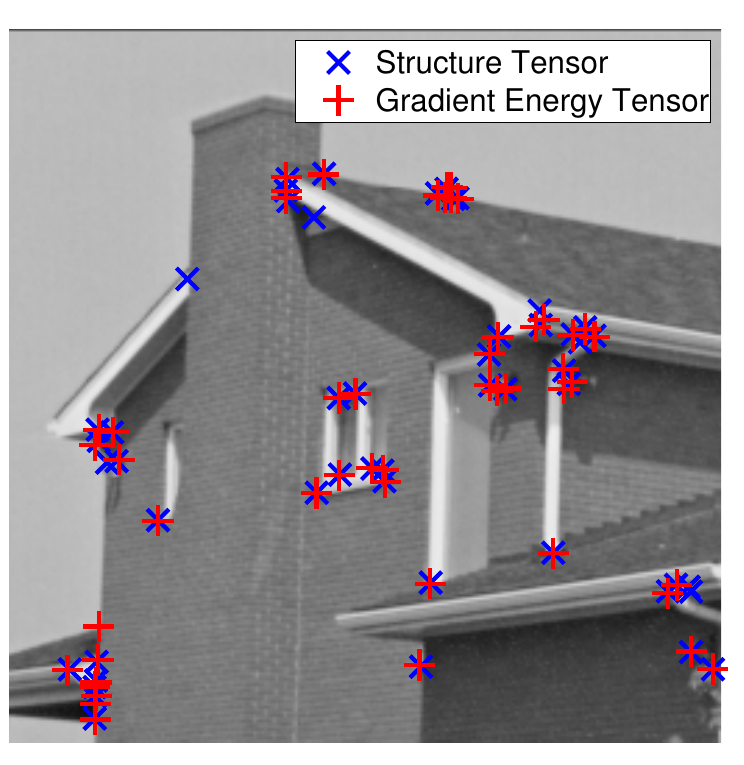} \\
		\hfill Corner Detection \hfill \phantom{I}\\
		\hfill \emph{Good Features to Track}~\cite{323794} \hfill \phantom{I}\\
	\end{minipage}
	\begin{minipage}{0.59\textwidth}
		\includegraphics[width=0.32\textwidth]{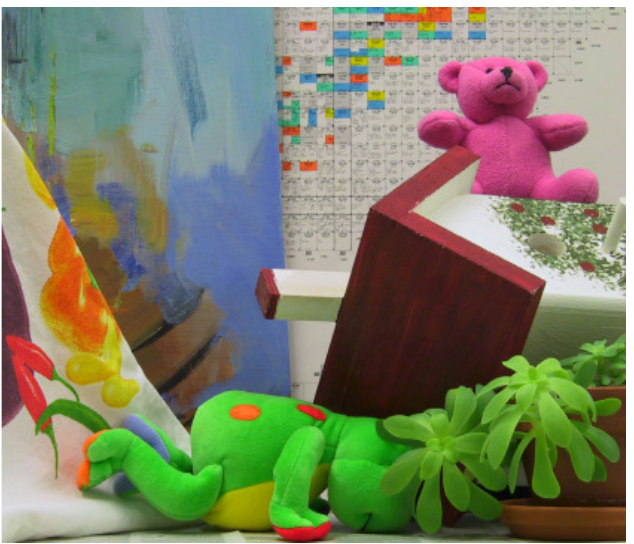}
		\includegraphics[width=0.32\textwidth]{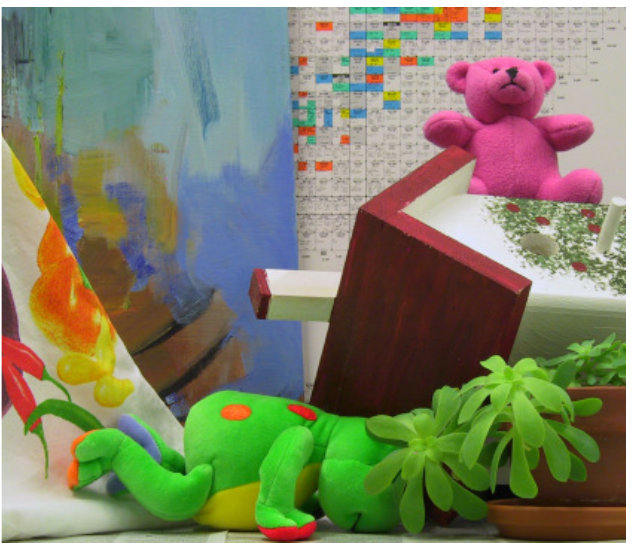}
		\includegraphics[width=0.32\textwidth]{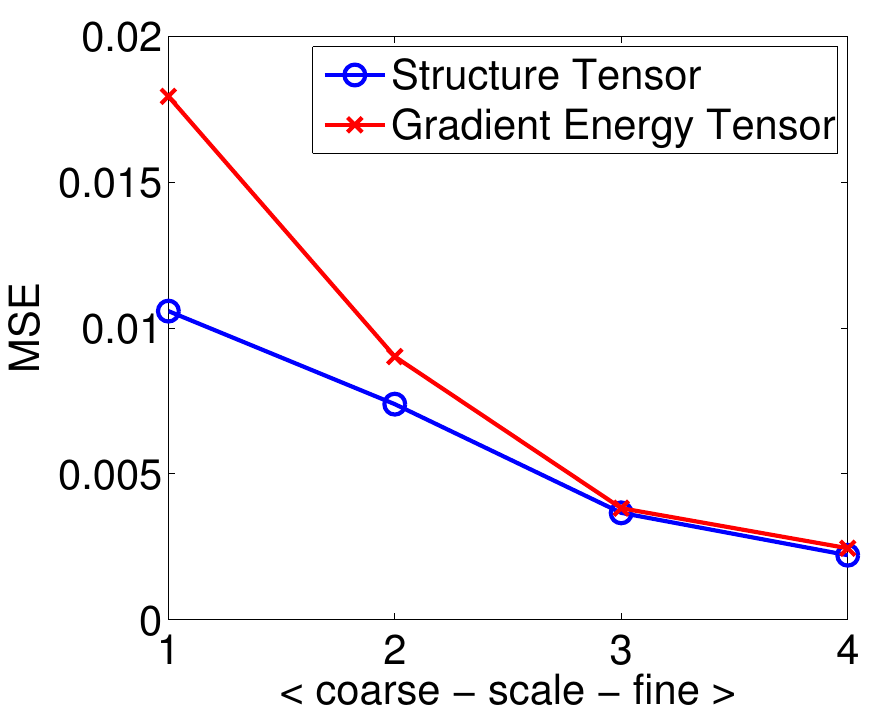} \\
		\includegraphics[width=0.48\textwidth]{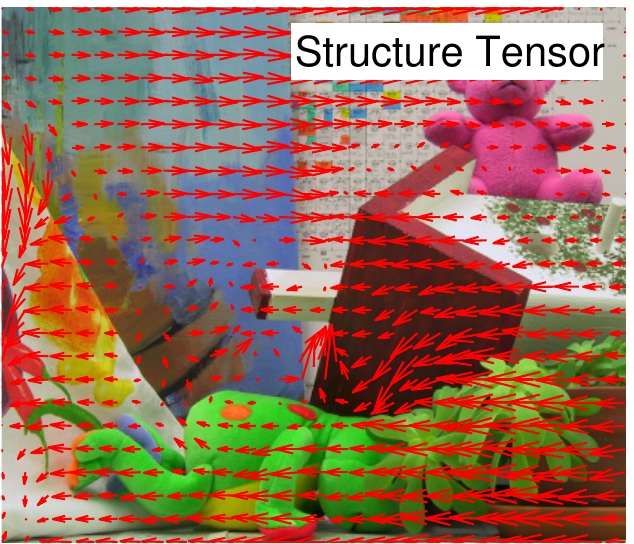}
		\includegraphics[width=0.48\textwidth]{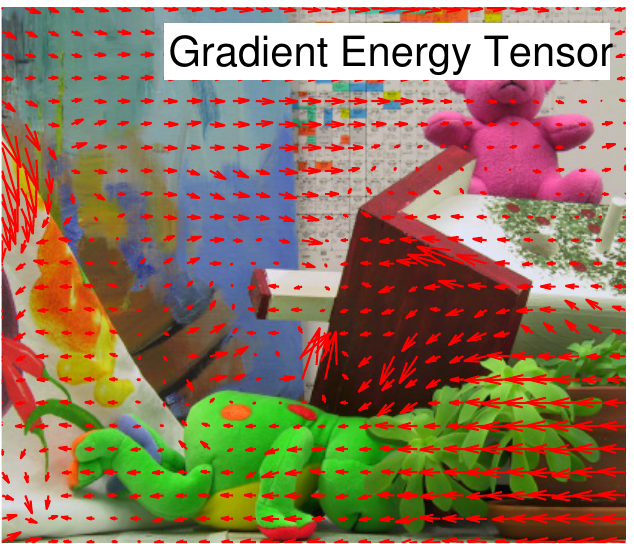} \\
		\hfill Optical flow \hfill \phantom{I}\\
		\hfill \emph{Dense Lucas-Kanade}~\cite{Lucas:1981:IIR:1623264.1623280} \hfill \phantom{I}\\
	\end{minipage} 
	\caption{An illustration that the Gradient Energy Tensor performs comparably to the Structure tensor in the two applications of corner detection and optical flow estimation (sequence Teddy frames 10 and 11~\cite{Baker:2011:DEM:1960926.1960983}). The gradient energy tensor does \textbf{\emph{not}} (necessarily) require a post-convolution of its tensor-components, which is the case in the case of corner detection. See text for details. }
	\label{fig:intro}
\end{figure*}

In this paper we propose to replace the structure tensor with an alternative tensor, the gradient energy tensor~\cite{diva2:269100} that does not (necessarily) require a post-convolution of its tensor-components to form a rank-2 tensor. The principal difference to the structure tensor is that the gradient energy tensor use higher-order derivatives to capture the orientation in a neighbourhood. In figure~\ref{fig:intro} we have used the tensors for corner detection and dense optical flow and as visualized the two tensors produce very similar results. As a major contribution of this work we present an adaptive tensor-based image denoising method implemented using Nvidia CUDA programming language. Our approach is superior to those based on the structure tensor with regards to computational performance, without loss of image quality.  

The structure tensor is defined as the outer product of the image gradient directions and in the case of two-dimensional images, the tensor is at most of rank-2 and determines the local energy distribution of a neighbourhood. In practice, to enforce robustness to noise and to form a rank-2 tensor, the structure tensor components are averaged using a post-smoothing of the tensor-components. Furthermore, without the post-convolution operation the tenor is of rank-1 and thus it cannot be used to describe corners and junctions in the image structure. The averaged structure tensor can also be viewed as a second-moment matrix, which estimates the local variance of the image data~\cite{lindeberg1993scale}. 

The gradient energy tensor is particularly suitable when considering graphical processing units (GPU). A GPU is a high-performance graphics card and is designed for massive parallelization of data-processing tasks. The GPU architecture is most suitable for problems with high spatial locality in the image plane and is therefore very fitting for the solution of partial differential equations (PDE). In this work we present a novel tensor-based PDE, which uses the gradient energy tensor for image enhancement and utilize the GPU architecture to enable real-time image enhancement. 

The standard approach to image filtering using PDE's is defined using the structure tensor, a rank-2 tensor which is transformed such that it describes the direction parallel to the image structure. This approach does not allow for real-time computations of high-resolution colour images since the tensor is defined using several convolutions of the tensor components. In contrast, the gradient energy tensor does not requires any convolution to form a rank-2 tensor but yet performs equally well, or better in denoising, when considering the resulting peak signal to noise ratio (PSNR) and structural similarity index (SSIM)~\cite{Wang2004}. 

\subsection{Contributions}
In this work we present a lesser known tensor: the gradient energy tensor as an alternative to the commonly used structure tensor. Our main contribution is a novel PDE-based denoising scheme which utilize the gradient energy tensor to drive an image enhancement process. In section~\ref{sec:corner_detection} we adopt the Good Features to Track~\cite{323794} framework to our tensor formulation and show comparable performance to the structure tensor using a repeatability test for different viewpoint angles. We also demonstrate that it is possible to solve the Lucas-Kanade~\cite{Lucas:1981:IIR:1623264.1623280} optical flow formulation using the gradient energy tensor. In section~\ref{sec:optical_flow} we illustrate the approach on sequences from the Middleburry optical flow dataset~\cite{Baker:2011:DEM:1960926.1960983}. 

Finally, in section~\ref{sec:denoising} we do an exhaustive evaluation on high-resolution colour images and describe the GPU implementaion using Nvidias CUDA programming language. We show that by using the gradient energy tensor we significantly boost the achieved frames per second (fps) compared to the structure tensor to enable real-time image denoising. 



\section{Estimating directional information}
Many image processing algorithms contain an estimate of local orientation as an integral part of the methods. Often the directional information is computed using the so called structure tensor \cite{Forstner87,diva2:274026}. The tensor is the outer product of the image gradients whereafter the components are averaged in a local neighbourhood. If the averaging operator, $w(x)$ is circular (\ie Gaussian) then the structure tensor is isotropic in homogeneous regions. Below, we define the structure tensor and the gradient energy tensor which we show in the experimental part outperforms the structure tensor with regards to computation efficiency without compromising the accuracy of the final result. 

\subsection{Structure tensor}
The structure tensor, $T \in \mathbb{R}^{2\times2}$, is symmetric and positive semi-definite~\cite{Forstner87,diva2:274026}. The tensor is defined as the outer product of the image gradients followed by post-convolutions, one for each component, \ie 
\begin{equation}\label{eq:ST}
	T(u_x, u_y) = \begin{pmatrix}
					{\displaystyle \int} w(\xi) u_x(x-\xi)^2 d\xi 			   & {\displaystyle \int} w(\xi) u_x(x-\xi)u_y(x-\xi) d\xi \\
					{\displaystyle \int} w(\xi) u_x(x-\xi)u_y(x-\xi) d\xi  & {\displaystyle \int} w(\xi) u_y(x-\xi)^2 d\xi
				\end{pmatrix}
\end{equation}
The effect of the post-convolution operator is that $T$ will have two non-zero eigenvalues, thus the operator can be used to estimate the orientation of image structures. In the case the convolution is given by the identity \ie $w(x) = 1$ for $x=0$ and $w(x) = 0$ otherwise, then the eigenvalues are given by the trace of T and $0$. 

\subsection{Gradient energy tensor}
\begin{figure}[t]
   \centering
	\hfill
	\includegraphics[width=0.49\textwidth]{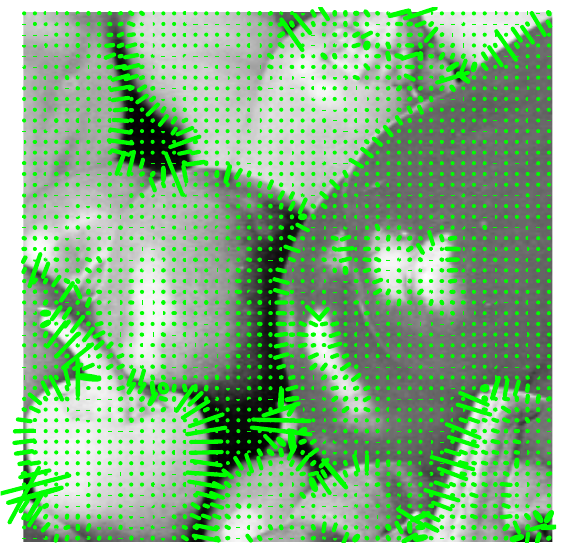}
	\hfill \hfill
	\includegraphics[width=0.49\textwidth]{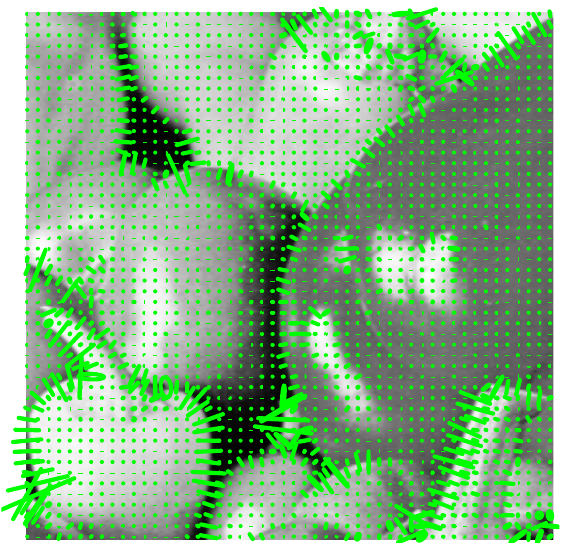} \hfill	\\
	\hfill \emph{Structure Tensor~\eqref{eq:ST}} \hfill \hfill \emph{Gradient Energy Tensor~\eqref{eq:GET}} \hfill \phantom{I}\\
	\caption{Illustration of the resulting tensor-fields. The structure tensor require additional post-smoothing to form a rank-2 tensor compared to the gradient energy tensor which is defined without post-smoothing. Note that the size of the ellipses has been scaled for improved visualization.}
	\label{fig:tensor_vis}
\end{figure}

The gradient energy tensor (GET) was first introduced by Felsberg and K{\"o}the \cite{diva2:269100}. Let $Hu = \nabla \nabla^t u$ be the Hessian and $\nabla\Delta u = \nabla \nabla^t \nabla u$, then GET is
\begin{equation}\label{eq:GET}
	GET(\nabla u) = HuHu - \frac{1}{2}\Big(\nabla u [\nabla \Delta u]^t + [\nabla \Delta u]\nabla^t u \Big)
\end{equation} 
In contrast to the structure tensor~\eqref{eq:ST}, the gradient energy tensor~\eqref{eq:GET} does \emph{not} (necessarily) require a convolution operator to form a rank 2 tensor. The response from the two tensors are illustrated in figure~\ref{fig:tensor_vis}. The presence of second and third-order derivatives in GET does makes it more sensitive to noise than the structure tensor, however, it allows us to capture orientation of structures not possible to detect using the structure tensor. 

The second difference to the structure tensor is that the gradient energy tensor is not necessarily positive semi-definite. In applications where it is required to have a positive semi-definite tensor it is straightforward to define the eigenvalues of GET to be positive. That is, simply compute the eigenvaluedecomposition of GET and use the alternative definition
\begin{equation}\label{eq:GET+}
	GET^+(\nabla u) = vv^t|\mu_1| + ww^t|\mu_2|
\end{equation}
where $v$, $w$ are the eigenvectors and $\mu_1$, $\mu_2$ eigenvalues of GET respectively. From~\eqref{eq:GET+} the tensor's orientation information is made explicit, the eigenvectors describe the local orientation of the neighbourhood and the eigenvalues describe the magnitude.  

Since the positivity of the tensor is reflected in the sign of its eigenvalues the presence of negative eigenvalues can be determined on beforehand if the condition 
	$\tr ( HuHu ) - \nabla^t u \nabla \Delta u \geq \sqrt{l}$
is not satisfied
where $l = (\tr{GET})^2 - 4\det(GET) \geq 0 $. Since GET is symmetric it has real eigenvalues. Thus by its eigenvalue decomposition it is sufficient to show that $\tr{GET} \geq \sqrt{l}$ in order for GET to be positive semi-definite. $l$ is necessarily positive since $l = (a-c)^2 + 4b^2 \geq 0$ where $a$, $b$ and $c$ are the components $GET(\nabla u) = \begin{pmatrix} a & b \\ b & c\end{pmatrix}$ and 
\begin{align}
\label{eq:a}	a &= u_{xx}^2 + u_{xy}^2 - u_x (u_{xxx} + u_{xyy}) \\
\label{eq:b}	b &= u_{xx}u_{xy} + u_{yx}u_{yy} - \frac{1}{2}(u_x (u_{yxx} + u_{yyy} )  + u_y(u_{xxx} + u_{xyy}) ) \\
\label{eq:c}	c &= u_{yy}^2 + u_{yx}^2 - u_y (u_{yxx} + u_{yyy}) 
\end{align}
An analysis of the positivity of the 1-dimensional energy operator was done in~\cite{275632}. In the remainder of the paper we apply the GET for corner detection, more specifically the good features to track approach, and $GET^+$ is used to compute the Lucas-Kanade optical flow and tensor-based image denoising. 




\subsection{Eigendecomposition of $2 \times 2$ tensors}
In the previous section we have shown that both the structure and gradient energy tensor can be used to compute the local orientation. By factorizing the tensors into their eigendecomposition the directional change and its magnitude is made explicit. Specifically, a matrix $S\in \mathbb{R}^{2\times 2}$ can be decomposed into its eigenvalues $\mu_{1,2}$ and eigenvectors $v, w$ representation such that 
\begin{equation}\label{eq:S_eigendecomp}
	S = vv^t\mu_1 + ww^t\mu_2
\end{equation}
where $v,w$ are two orthonormal vectors. The eigenvectors describe the orientation and the eigenvalues the magnitude of the directional change in a neighbourhood. The eigenvalues can be computed by solving the characteristic polynomial $\det(S - \mu I) = 0$ and the solution is given by
\begin{equation}\label{eq:S_eig}
	\mu_{1,2} = \frac{1}{2} \left ( \tr{S} \pm \sqrt{(\tr S)^2 - 4\det{S}} \right )
\end{equation}
For the applications presented in this work we are not required to compute the explicit eigendecompistion~\eqref{eq:S_eigendecomp}, rather we are primarily interested in the eigenvalues. Thus, by observing that $vv^t + ww^t = I \iff ww^t = I - vv^t$, then $S$ can be expressed as~\cite{books/daglib/0095066}, 
\begin{equation}\label{eq:S_eigdecomp}
	S = (\mu_1 - \mu_2)vv^t + I\mu_2
\end{equation}
and we compute the eigenvector-product $vv^t$ as
\begin{equation}
	vv^t  = \frac{1}{(\mu_1 - \mu_2)}(S - I\mu_2)
\end{equation}

\section{GET corner detection and optical flow}

\begin{figure*}[t]
   \centering
   \hfill
   	\includegraphics[width=0.40\textwidth]{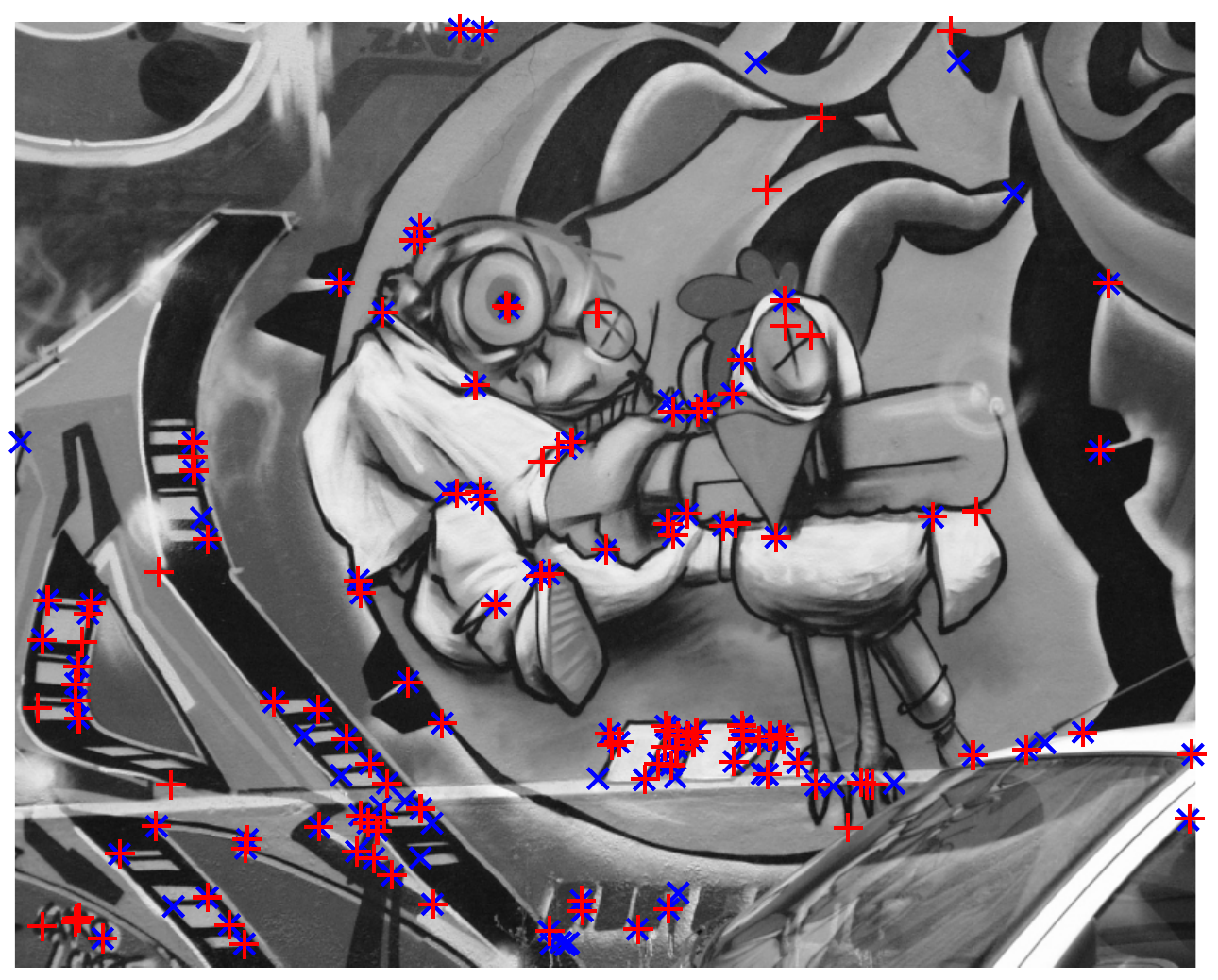}
	\hfill \hfill
	\includegraphics[width=0.40\textwidth]{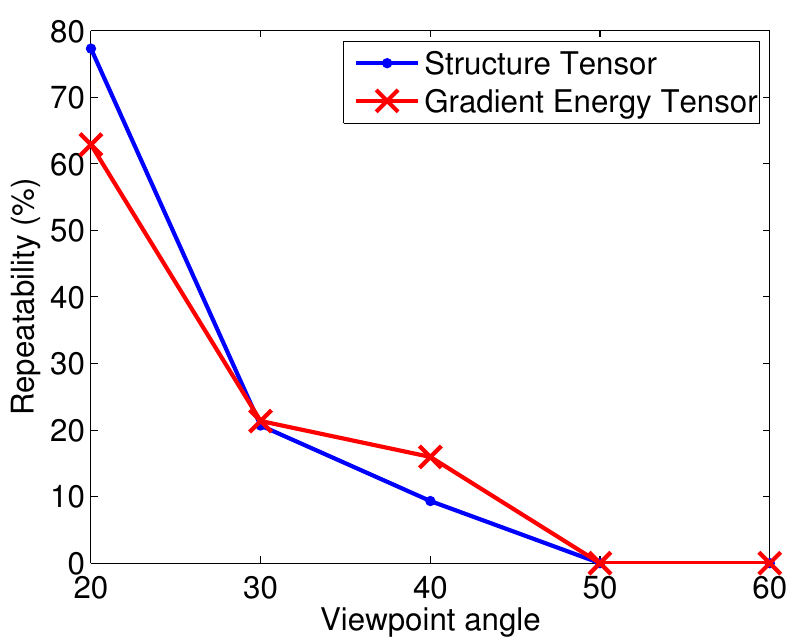}
	\hfill \phantom{I} \\
	\hfill	Reference \hfill \hfill \hfill \phantom{I} \\[3mm]
	\hfill
	\includegraphics[width=0.40\textwidth]{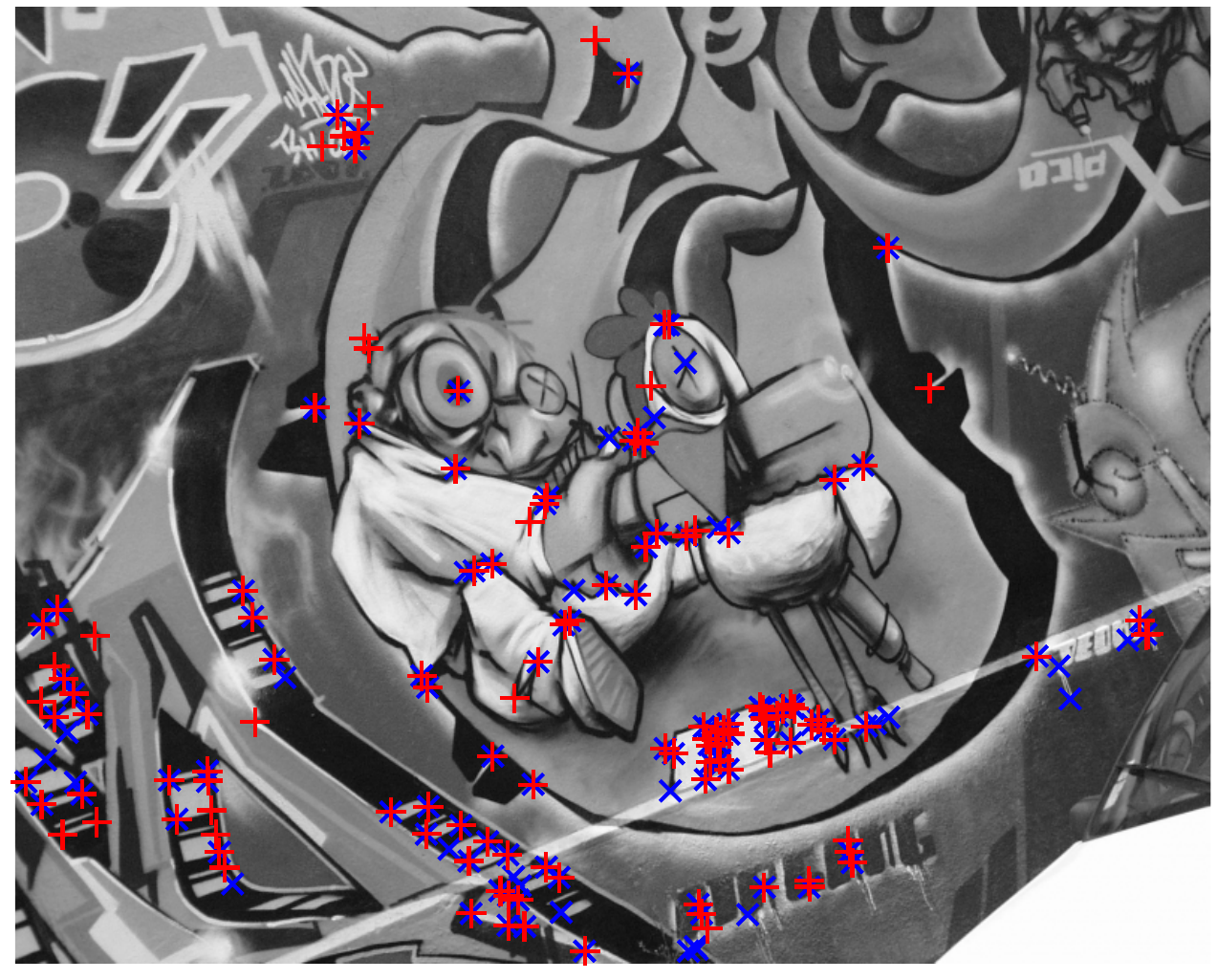}
	\hfill \hfill
	\includegraphics[width=0.40\textwidth]{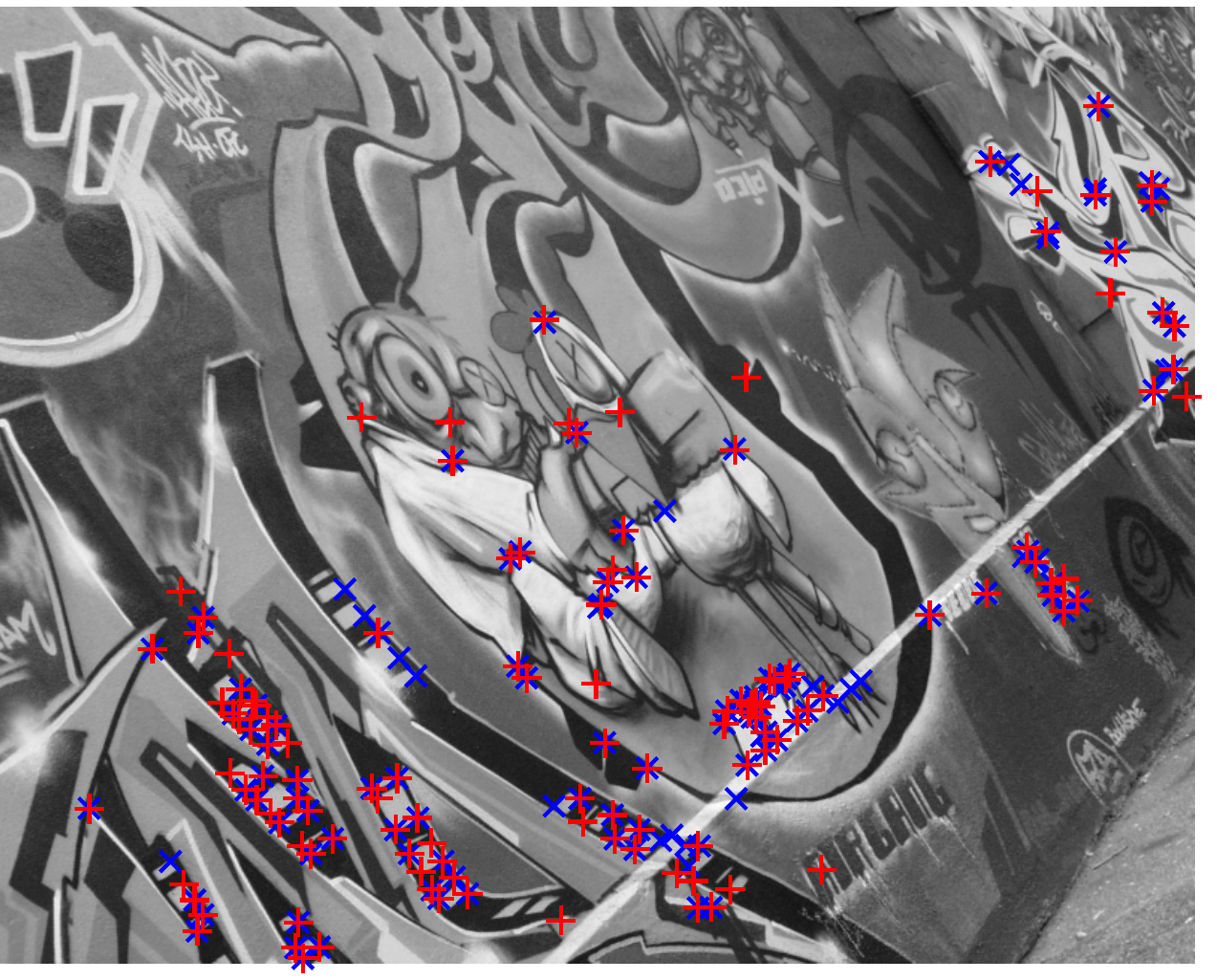}
	\hfill \phantom{I} \\
	\hfill	20 degrees viewpoint angle \hfill \hfill 40 degrees viewpoint angle \hfill \phantom{I} \\
	\caption{Examples of corner detection for the structure tensor (with post-smoothing) and the gradient energy tensor (without post-smoothing) using Good Features to Track. Both tensors detect corners accurately but not always the same corners.}
	\label{fig:corner_detection_repeatability}
\end{figure*}

\subsection{Corner detection}
\label{sec:corner_detection}
The first application we consider is corner detection using Good Features to Track~\cite{323794}. The problem of corner detection is part of many image processing pipelines such as interest point detection and sparse optical flow. The Good Features to Track framework detects corners by considering the eigenvalues of the structure tensor. If the structure tensor has two non-zero eigenvalues $\mu_{1,2}$, both larger than some threshold $\mu$ then the orientation tensor is necessarily invertible, \ie the tensor is of rank 2. If $\min(\mu_1,\mu_2) > \mu$
where $\mu$ is often set to a fraction of the largest minimum eigenvalue then the neighbourhood contains a corner point. 

In this work we set $\mu = 0.01$ and figure~\ref{fig:corner_detection_repeatability} shows the 128 strongest detected corners for the structure tensor and the gradient energy tensor. We use a Gaussian kernel with standard deviation 1 for post-smoothing of the structure tensor components. We also computed the repeatability measure~\cite{1498756} at 40\% overlap (see figure~\ref{fig:corner_detection_repeatability}) for the \emph{viewpoint} dataset where the viewpoint angle has been changed from 20-60 degrees from the reference image~\cite{repeatability2014}. The repeatability measure is similar for the two tensors. Note that the gradient energy tensor, in this example, does not contain a post-smoothing of the tensor components. 





\begin{figure*}[t]
   \centering
		\hfill
		\includegraphics[width=0.49\textwidth]{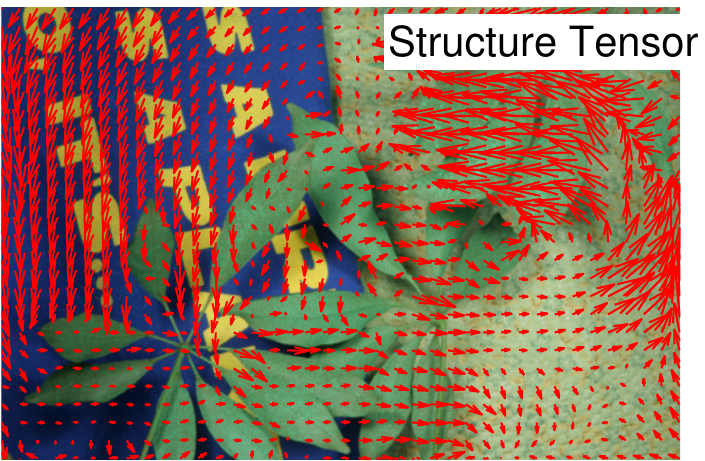}
		\hfill \hfill
		\includegraphics[width=0.49\textwidth]{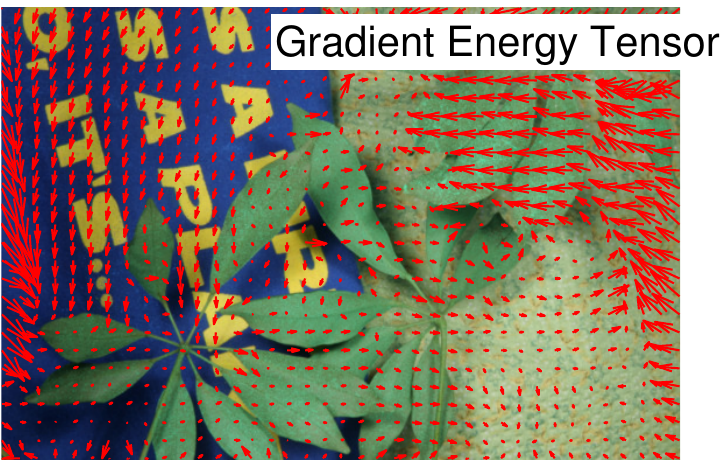} \hfill \phantom{I}\\
		\begin{minipage}{0.75\textwidth}
			\hfill Flow fields \hfill \phantom{I}\\
			\hfill \hspace*{0mm} \emph{Structure Tensor} \hfill \hfill \hspace*{0mm} \emph{Gradient Energy Tensor} \hfill \phantom{I}\\[-2mm]
			\hfill
			\hspace*{-5mm}
			\includegraphics[width=0.49\textwidth]{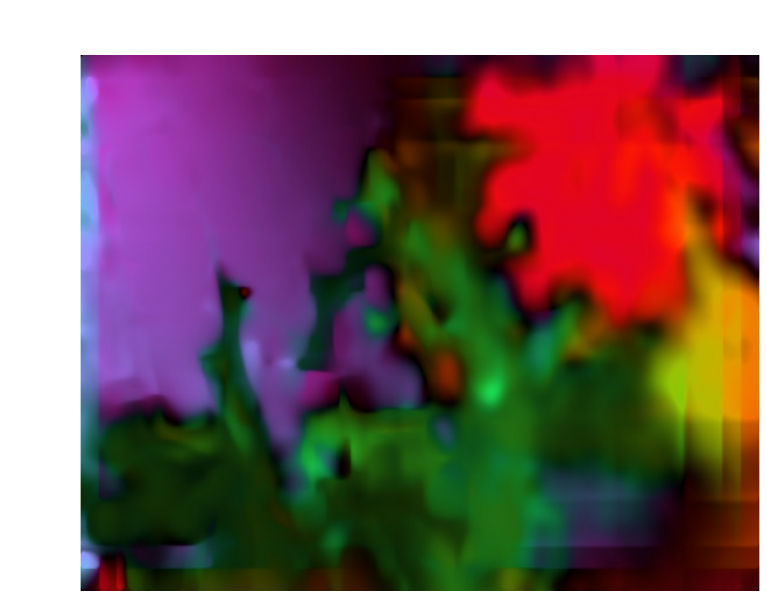}
			\includegraphics[width=0.49\textwidth]{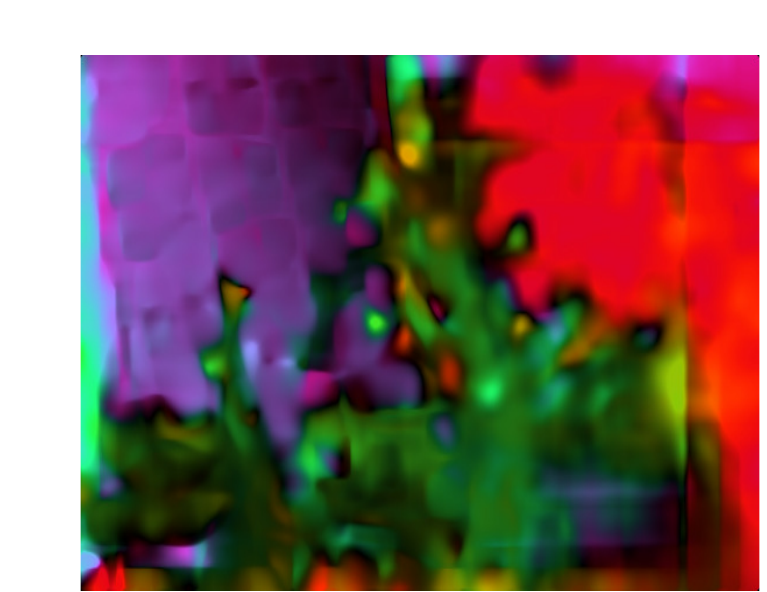}
			\hfill \phantom{I} \\
		\end{minipage}
		\begin{minipage}{0.24\textwidth}
			\hfill
			\vspace*{5mm}
			\hspace*{-5mm}
			\includegraphics[width=1.2\textwidth]{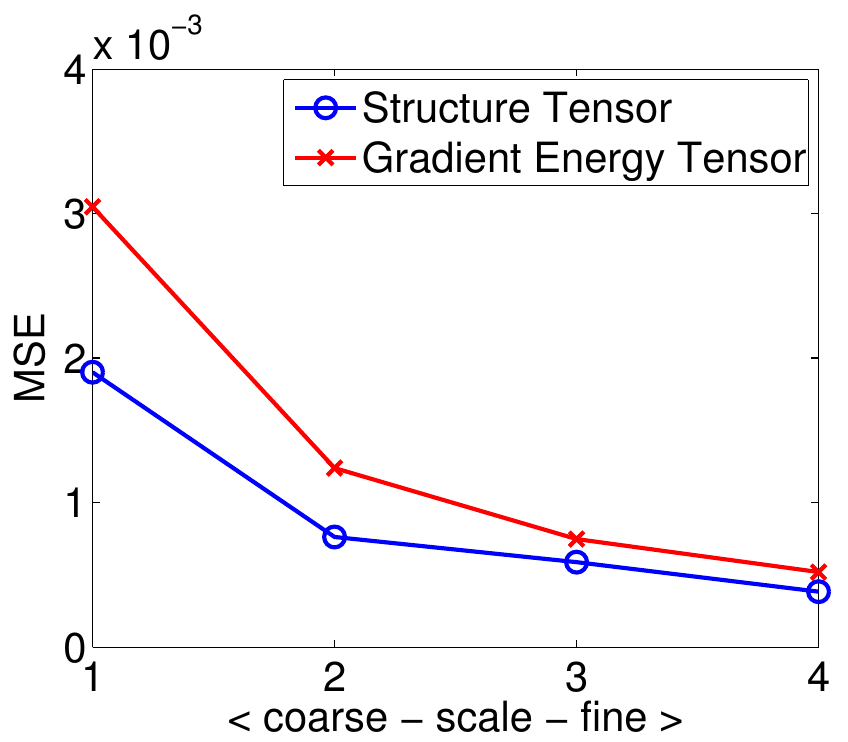} 
			\hfill \phantom{I} \\
		\end{minipage}
	\caption{Optical flow estimated from frames 7 and 8 of the Schefflera sequence in the Middleburry optical flow dataset~\cite{Baker:2011:DEM:1960926.1960983}. Top row illustrate the vector field. On the bottom we show the obtained flow fields where the direction is colour coded and intensity corresponds to the magnitude. The graph to the right show the mean squared error between the warped image $J(x+d)$ and the reference image $I(x)$ in~\eqref{eq:E_OF}.}
	\label{fig:optical_flow}
\end{figure*}

\begin{figure}[t]
	\centering
	\hfill
	\includegraphics[width=0.24\textwidth]{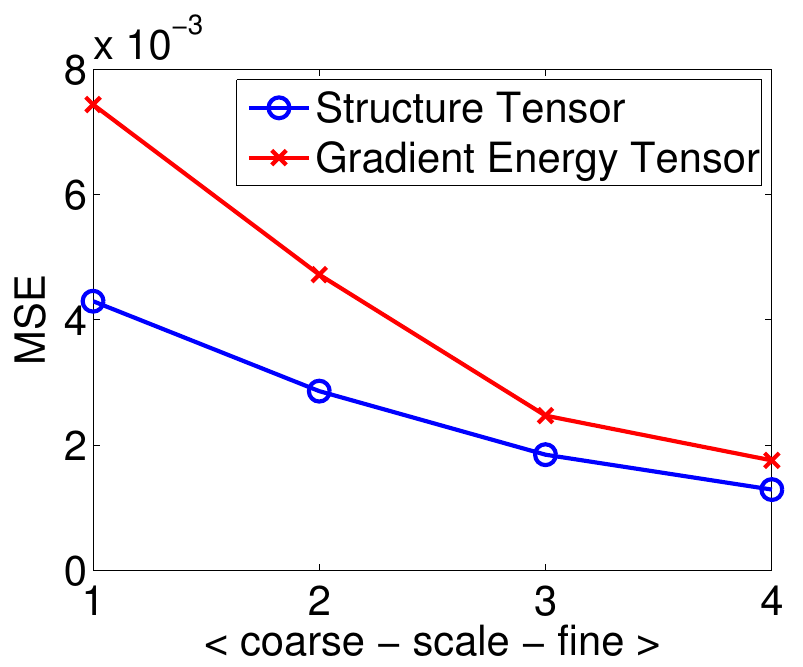} 
	\hfill\hfill
	\includegraphics[width=0.24\textwidth]{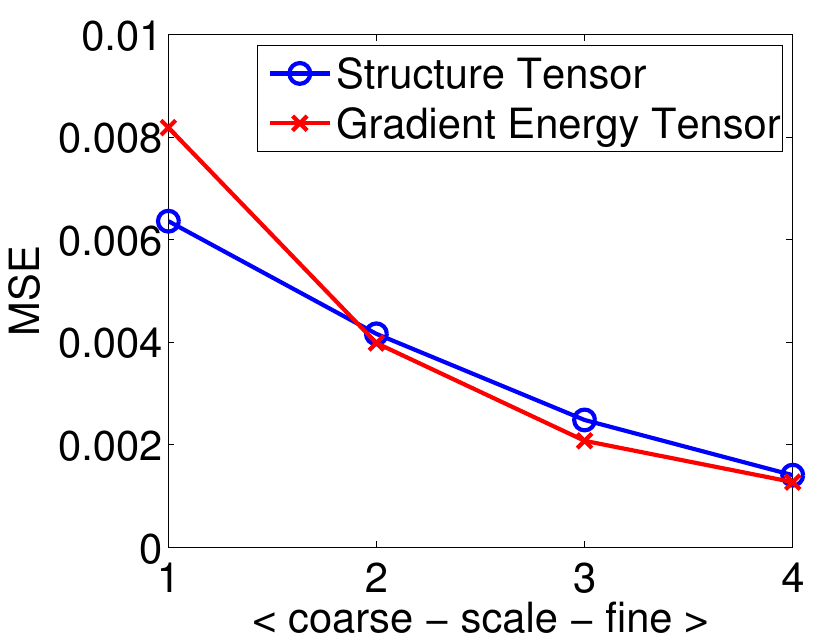} 
	\hfill\hfill
	\includegraphics[width=0.24\textwidth]{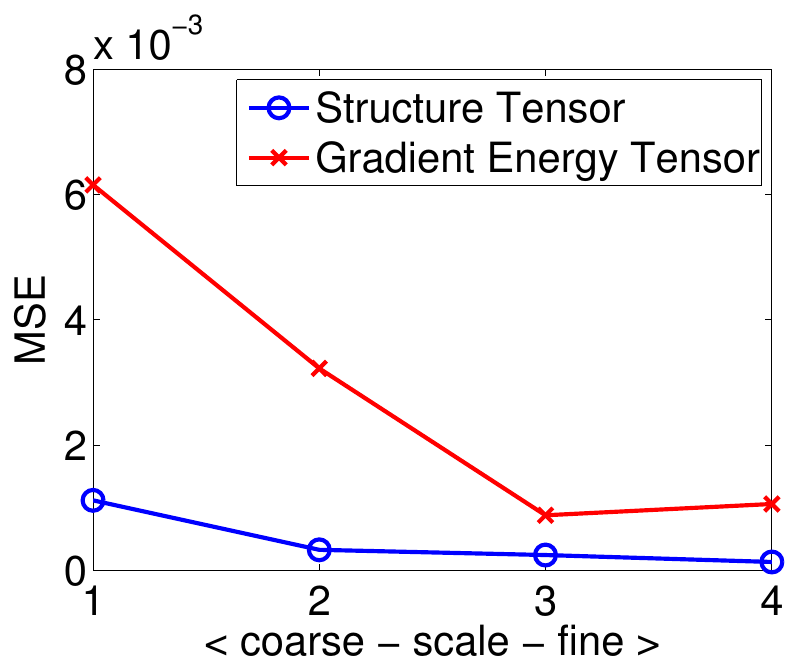} 
	\hfill\hfill
	\includegraphics[width=0.24\textwidth]{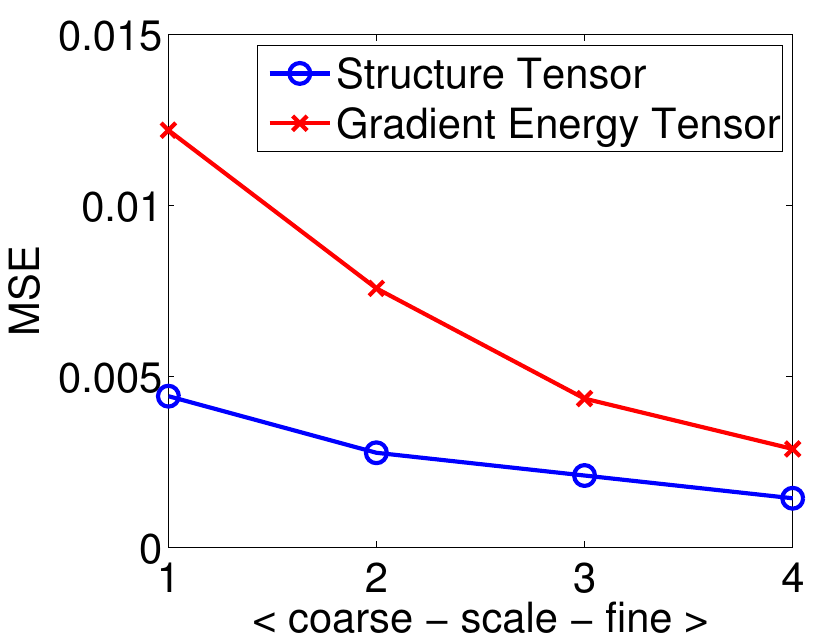} \hfill \\
	\hfill Backyard \hfill \hfill  Dumptruck \hfill \hfill Wooden \hfill \hfill  Yosemite \hfill  \phantom{I} \\	
	\caption{Mean squared error (MSE) graphs between the first and second image after warping the first image with the estimated motion field. The sequences are part of the Middlebury dataset~\cite{Baker:2011:DEM:1960926.1960983}. The estimate of the motion field in the Wooden sequence for the gradient energy tensor diverge on the finest scale, but in the other sequences the MSE yield comparable final displacement estimate.}
	\label{fig:optical_flow_mse}
\end{figure}

\subsection{Optical flow}
\label{sec:optical_flow}
Our second application is to apply the gradient energy tensor to the original Lucas-Kanade optical flow formulation~\cite{Lucas:1981:IIR:1623264.1623280} to compute a dense motion field. By minimizing the below energy functional the structure tensor appear as part of the minimizer
\begin{equation}\label{eq:E_OF}
	E(u) = \int_\Omega [J(x + d) - I(x)]^2 w(x)  \; dx
\end{equation}
where $J$ and $I$ are two images of size $\Omega$ with some unknown displacement vector $d$. The minimizer of~\eqref{eq:E_OF} is obtained by solving (see~\cite{Tomasi1991})
\begin{equation}\label{eq:LKmin}
	\left [ \int_\Omega [\nabla J(x) \nabla^t J(x)] w(x) \; dx \right ] d  = \left [ \int_\Omega [(J(x) - I(x))\nabla J(x)] w(x) \; dx \right ]
\end{equation}
The bracket in the left hand side of~\eqref{eq:LKmin} is the structure tensor, $T$ in~\eqref{eq:ST}. Figure~\ref{fig:optical_flow} shows the results when we solve~\eqref{eq:LKmin} with the structure tensor and when we interchanged the structure tensor with the gradient energy tensor with positive eigenvalues, \ie $GET^+$ in \eqref{eq:GET+}. Due to the large displacement between the image pairs we are required to have a post-convolution of the $GET^+$ components similarly to the structure tensor in order to capture the motion. 

For both tensors we solve the normal equation~\eqref{eq:LKmin} explicitly using the pseudo-inverse on multiple scales. The solution we obtain at a coarse scale is propagated to a finer scale and after each scale we apply a median-filter to the estimated motion field~\cite{5539939}. Furthermore, we found that the magnitude scaling of the gradient energy tensor eigenvalues resulted in a poor motion estimation. Therefore, we scale the eigenvalues of the gradient energy tensor using the factor $\sigma i/I$ where $i \in I = \{1,2,3,4\}$ are the scales, $i=1$ is the fines scale, and $\sigma$ is the standard deviation of the Gaussian filter $w(x)$. As for the structure tensor, the selection of $\sigma$ is dependent on the absolute motion present within the frames, \ie if the displacement is large then a larger $\sigma$ is required. In figure~\ref{fig:optical_flow_mse} we show the mean squared error between $J(x+d)$ and $I(x)$ for some additional sequences from the Middlebury optical flow dataset~\cite{Baker:2011:DEM:1960926.1960983}. 

The optical flow formulation~\eqref{eq:E_OF} does (obviously) not give state of the art results, however the approach illustrates that the gradient energy tensor is a possible alternative to the structure tensor. We expect that many interesting further results can be derived from this approach. 



\section{Iterative tensor-based PDE denoising}
\label{sec:denoising}
Image enhancement methods based on partial differential equations (PDE) are often considered to be too computationally expensive for practical applications. The main bulk of computations is the iterative update scheme and the calculation of the post-convolution of the structure tensor components~\cite{Weickert96anisotropicdiffusion}. 

It is in this application that the gradient energy tensor really excels over the structure tensor. In this section we implement the proposed iterative denoising scheme on the GPU. We show how to utilize the highly parallelizable nature of iterative PDE-based denoising schemes and benefit from the locality of the gradient energy tensor. The implementation is done using Nvidias CUDA programming language with OpenGL support, the GPU we use is the GTX 670 with 4Gb on card memory and the workstation is equipped with an Intel Xeon CPU at 3 GHz and 8 Gb of installed RAM memory. Even though the hardware specification is in the middle segment of the consumer-market we show that by using the gradient energy tensor, the proposed iterative tensor-based PDE denoising scheme can reach near maximum PSNR at 60 frames per second (fps) for a three channel colour $1280 \times 720$ pixel image (HD720p). This is a significant improvement over the structure tensor running at 30 fps while achieving similar peak signal to noise ration (PSNR) and structural similarity~\cite{275632} (SSIM) values.  

\subsection{The proposed filtering scheme}
The standard formulation of tensor-based anisotropic diffusion~\cite{Weickert96anisotropicdiffusion} using the structure tensor is given in the PDE below with Neumann boundary condition
\begin{equation}\label{eq:PDE-T}
   \hspace{0mm}\left \{
    \begin{array}{rl}
     u-u^0 - \beta  \div{ D(\nabla u) \nabla u }  & \hspace*{0mm} = 0 \mbox{ in }  \Omega \\
      {\bm n} \cdot \nabla u & \hspace*{0mm} =  0     \mbox{ on }  \partial \Omega 
     \end{array} \right .
\end{equation}
where $\beta$ is a stepsize parameter which controls the smoothness of the solution $u$ that minmizes the PDE. In~\eqref{eq:PDE-T}, $D(\nabla u)$ is the \emph{diffusion tensor} computed as
\begin{equation}\label{eq:D(T)}
	D(T(\nabla u)) = vv^t g(\mu_1) + ww^tg(\mu_2)
\end{equation}
where $v,w$ and $\mu_{1,2}$ are the eigenvectors and eigenvalues of the structure tensor $T$ in~\eqref{eq:ST}. $g$ is the diffusivity function and here we set it as $g(s) = \exp(-s/k)$ where $k$ is the edge-stopping parameter determining the adaptivity to the image structure. Instead of using the diffusion tensor in the PDE~\eqref{eq:PDE-T} we propose to use the gradient energy tensor with positive eigenvalues, $GET^+$ in~\eqref{eq:GET+}, as the tensor controlling the orientation estimate of the image structures, \ie we define the following PDE 
\begin{equation}\label{eq:PDE-GET+}
   \hspace{0mm}\left \{
    \begin{array}{rl}
     u-u^0 - \beta  \div{ D^+(\nabla u) \nabla u }  & \hspace*{0mm} = 0 \mbox{ in }  \Omega \\
      {\bm n} \cdot \nabla u & \hspace*{0mm} =  0     \mbox{ on }  \partial \Omega 
     \end{array} \right .
\end{equation}
The computation of the eigenvalues are done according~\eqref{eq:S_eigdecomp}, \ie
\begin{equation}\label{eq:D+}
	D^+(GET^+(\nabla u)) = \left ( \frac{g(\lambda_1) - g(\lambda_2)}{\lambda_1 - \lambda_2} \right ) (GET^+-I\lambda_2) + Ig(\lambda_2)
\end{equation}
where $\lambda_{1,2}$ are the eigenvalues of $GET^+$ computed according to~\eqref{eq:S_eig} where we set $S = GET^+$. In practice we compute~\eqref{eq:D(T)} using~\eqref{eq:D+} with $S=T$. 

In order to solve the PDEs~\eqref{eq:PDE-T} and~\eqref{eq:PDE-GET+}, we use a forward Euler iterative scheme with finite differences to approximate the image derivatives. For a discussion on the numerical stability of iterative scheme see~\cite{Scherzer2000}.


\subsection{Implementation details}
A GPU implementation is about how to efficiently utilize the parallelism of the graphics card architecture. Using CUDA terminology, the parallelism is achieved by dividing the image data into blocks, each block contains the threads that are to be executed in parallel on a streaming multiprocessor. Today's GPU architectures offer many memory types (global, texture, shared, local ...) and our implementation is focused on utilizing the high-performance shared memory. We achieve this by coalescing memory access when streaming data from global memory to shared memory. We considered using texture memory due to its automatic handling of Neumann boundary conditions and memory-access optimization for localized reads, however texture memory is read-only and we require dynamic updates of intermediate results. Also, shared memory is on-chip, and therefore read-access requires less clock-cycles than the texture memory. These differences in memory-latency have a significant impact on runtime performance, for example in convolutions~\cite{Podlozhnyuk2007}. 
\begin{table}[t]
\centering
\begin{tabular*}{\columnwidth}{@{\extracolsep{\stretch{1}}}*{7}{r}@{}}
	Anisotropic Diffusion~\cite{Weickert96anisotropicdiffusion} & \hspace{6mm} Gradient Energy Diffusion 
\end{tabular*}
\begin{minipage}{0.49\textwidth}
\begin{lstlisting}[frame=single,linewidth=5.6cm]  % Start your code-block

for i=0 to maxint do {
 convolution_rows()
 convolution_cols()
 compute_structure_tensor()
 filter_update()
}
\end{lstlisting}
\begin{lstlisting}[frame=single,linewidth=5.6cm]  % Start your code-block

compute_structure_tensor() {
 gradient_products(a,b,c)
 convolution_rows(a)
 convolution_cols(a)
 convolution_rows(b)
 convolution_cols(b)
 convolution_rows(c)
 convolution_cols(c)
 remapp_eigenvalues(a,b,c)
}
\end{lstlisting}
\end{minipage}
\hspace{5mm}
\begin{minipage}{0.49\textwidth}
\vspace{-15mm}
\begin{lstlisting}[frame=single,linewidth=5.6cm]  % Start your code-block

for i=0 to maxint do {
 convolution_rows()
 convolution_cols()
 compute_energy_tensor()
 filter_update()
}
\end{lstlisting}
\begin{lstlisting}[frame=single,linewidth=5.6cm]  % Start your code-block

compute_energy_tensor() {
 a = (4)
 b = (5)
 c = (6)
 remapp_eigenvalues(a,b,c)
}
\end{lstlisting}
\end{minipage}
\caption{Algorithm pseudo-code for the main CUDA kernels. Left: Standard approach to adaptive image diffusion using the structure tensor. Right: adaptive image diffusion using the gradient energy tensor. The convolutions are separable Gaussian functions of size $5 \times 5$ with standard deviation of $1$.}
\label{table:algorithm}
\end{table}

\begin{figure*}[t]\label{fig:shared_memory}
   \centering
	\hfill
	\raisebox{2mm}{
      \includegraphics[width=0.20\textwidth]{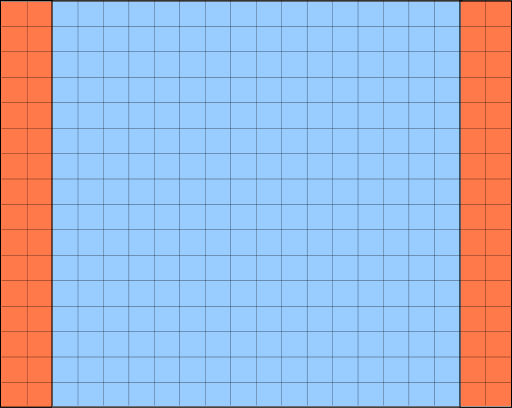}%
    }
	\hfill \hfill
	\includegraphics[width=0.15\textwidth]{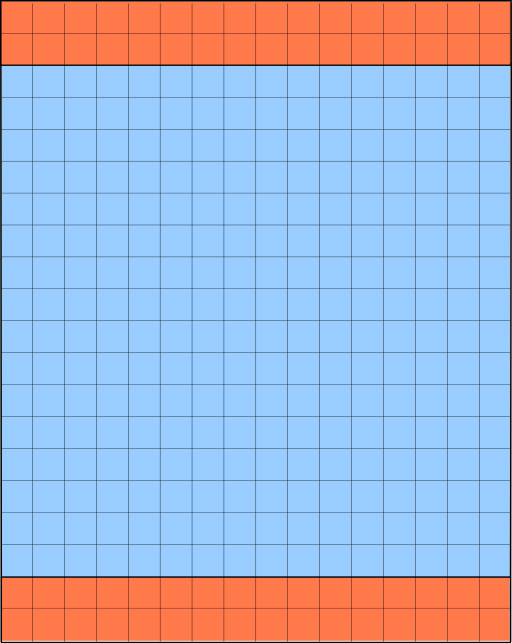}
	\hfill \hfill
	\includegraphics[width=0.20\textwidth]{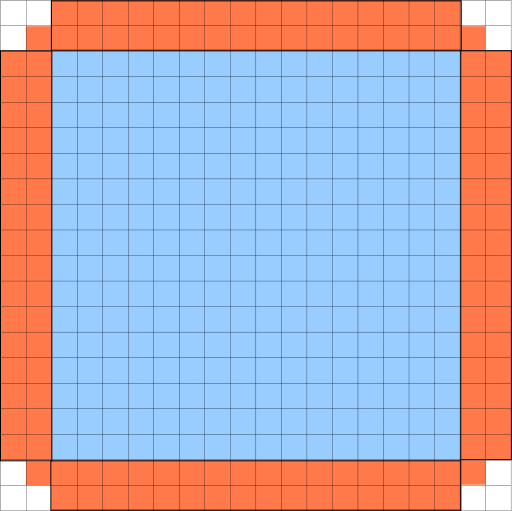}
	\hfill \phantom{I} \\ 
	\hfill
	(a) Convolution rows
	\hfill \hfill
	(b) Convolution columns
	\hfill \hfill
	(c) Gradient energy tensor
	\hfill \phantom{I} \\
	\caption{We use tiles of size $16 \times 16$ (blue regions) with padding of 2 pixels (red region) for the convolution and computation of the gradient energy tensor. We use corresponding tile layouts for computing the image gradient and filter update.}
	\label{fig:shared_memory}
\end{figure*}

Since we are interested in processing images with three colour channels, we simplify our implementation by defining a container describing the three colour channels red, green and blue as well as the alpha channel (required for visualization using OpenGL). The image data is read and written using 24-bits but during the filter process we use single precision float. There are primarily three steps involved in the iterative filtering scheme, pre-filtering for regularization of the image derivatives, orientation estimation and filter update. The steps are illustrated in table~\ref{table:algorithm} and highlights the primary difference between the two implementations: the computation of the structure tensor requires three full (separable) convolutions of the image data whereas the energy tensor does not require any post-convolution of the tensor-components, which is key to the gain in computation speed. Figure~\ref{fig:shared_memory} shows the memory layout of the shared memory that we use in the CUDA kernels shown in table~\ref{table:algorithm}. We use tile sizes of $16 \times 16$ and note that each entry in the tile consists of four entries \ie red, green, blue and alpha channels. This approach is convenient since it both simplifies the code and facilitates efficient memory access. The padding of the tiles (the red region in figure~\ref{fig:shared_memory}) is done with two pixels in the case of the separable convolution since we use a Gaussian filter of size $5 \times 5$ with standard deviation of 1 for smoothing the image and tensor-components. The coefficients of the Gaussian filters is set using constant memory. The shared memory associated with the gradient energy tensor require a padding of 2 pixels in horizontal and vertical direction since the support of the third order finite difference derivative is 5 pixels. Lastly, since we only require first order diagonal derivatives, it is sufficient to read the closest corner-pixel into the shared memory, further simplifying the global memory access pattern.

\begin{figure*}[t]
   \centering
	\includegraphics[width=1\textwidth]{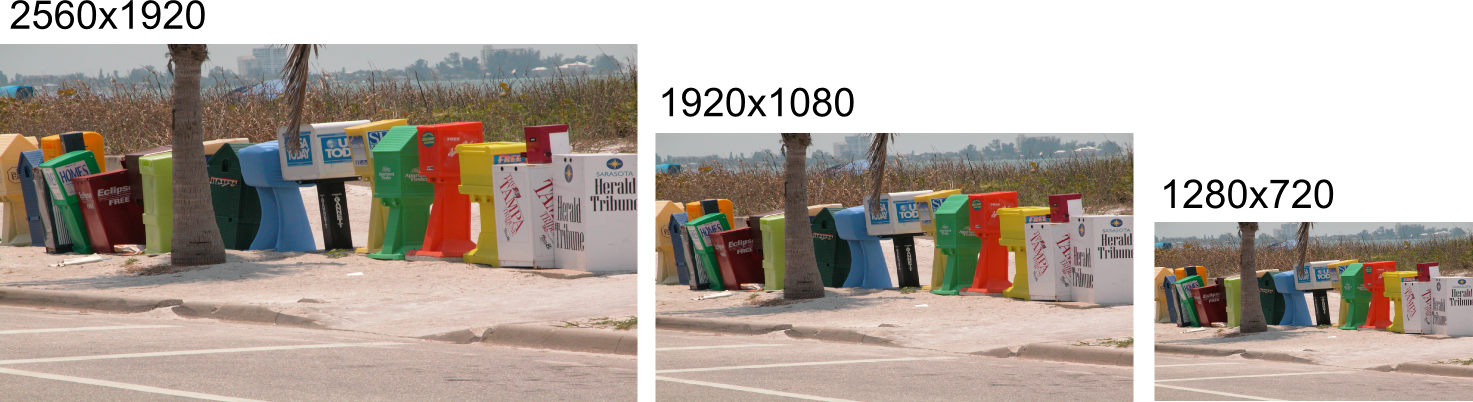} 
	\caption{Colour test images and the corresponding image sizes in pixels that are used in the evaluation.}
	\label{fig:test_images}
\end{figure*}

In order to measure the resulting computation times, we are required to compute the resulting execution speed of the filtering methods. We do this by using the standard sdkStartTimer() and sdkStopTimer() available in CUDA. We have chosen to include the OpenGL rendering in the timing of the filter performance as shown in table~\ref{table:cuda_timing} since it more accurately reflects the expected real-time capabilities of video denoising where each frame in a video sequence would be visualized. 
\begin{table}[t]
\begin{lstlisting}[frame=single]  % Start your code-block

void display() {
 sdkStartTimer(&timer);
 unsigned int *dResult;
 // Initialize OpenGL for visualization
 diffusionFilterRGBA(dResult, ...);
 // Map dResult to OpenGL resources and draw image on display
 sdkStopTimer(&timer);
}
\end{lstlisting}
\caption{Main function of the tensor-based image denoising method. The timer values reported in section~\ref{sec:results} are timed including the OpenGL rendering.}
\label{table:cuda_timing}
\end{table}


\subsection{Results}
\label{sec:results}
The focus of our evaluation is to show that the gradient energy tensor does not compromise the denoising quality compared to the structure tensor while achieving a faster runtime. Our measures include the peak signal to noise ratio (PSNR) and the structure similarity index~\cite{Wang2004} (SSIM) for the image quality. With regards to the total runtime (measured as shown in table~\ref{table:cuda_timing}) we report achieved frames per seconds (fps) for each method. For each of the measures a higher value is better than a lower value.

Figure~\ref{fig:test_images} show the colour test image \emph{pippin\_Florida0002.bmp} from the McGill colour image database~\cite{McGillDatabase} with the original resolution $2560\times 1920$ pixels. The image was downsampled (using bicubic interpolation) to $1920\times 1080$ (HD1080) and $1280\times 720$ (HD720), two common image resolutions in high-definition (HD) video. We corrupt each image with 20 standard deviations of normal distributed noise. The filtering scheme is iterated 10 times with a fixed update step of size 0.20 (we refer to~\cite{Scherzer2000} for a discussion on convergence results for iterated solutions of PDEs). The diffusivity constant (see $g$ in~\eqref{eq:D+}) is set to $k/10$ where we compute $k = (e^1-1)/(e^1-2)\sigma^2$~\cite{Fberg2011}, this yields $k/10=0.0015$ when $\sigma=20/255$ and each colour channel is quantized using an 8-bit representation. 

\renewcommand{\figwidth}{0.45}
\begin{figure*}[t]
   \centering
\includegraphics[width=\figwidth\textwidth]{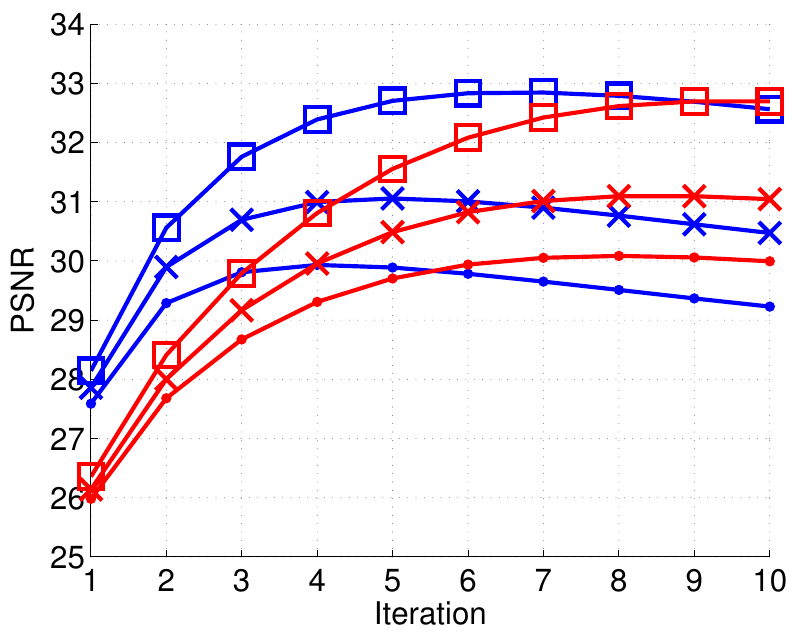}
	\hfill \hfill
	\includegraphics[width=\figwidth\textwidth]{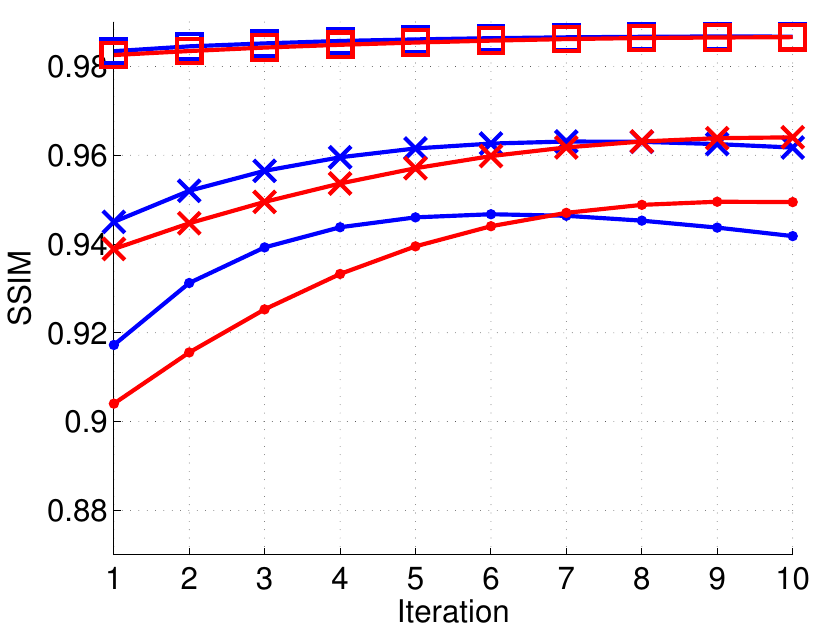}
	\hfill  \\
	\hfill (a) \hfill \hfill (b) \hfill \phantom{I} \\
   \hfill
	\includegraphics[width=\figwidth\textwidth]{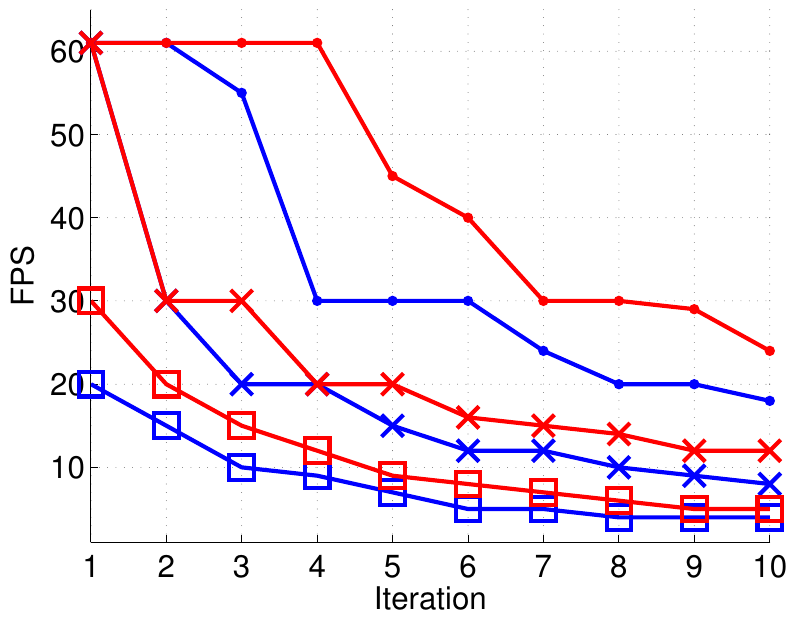} 
	\hfill \hfill
    \includegraphics[width=\figwidth\textwidth]{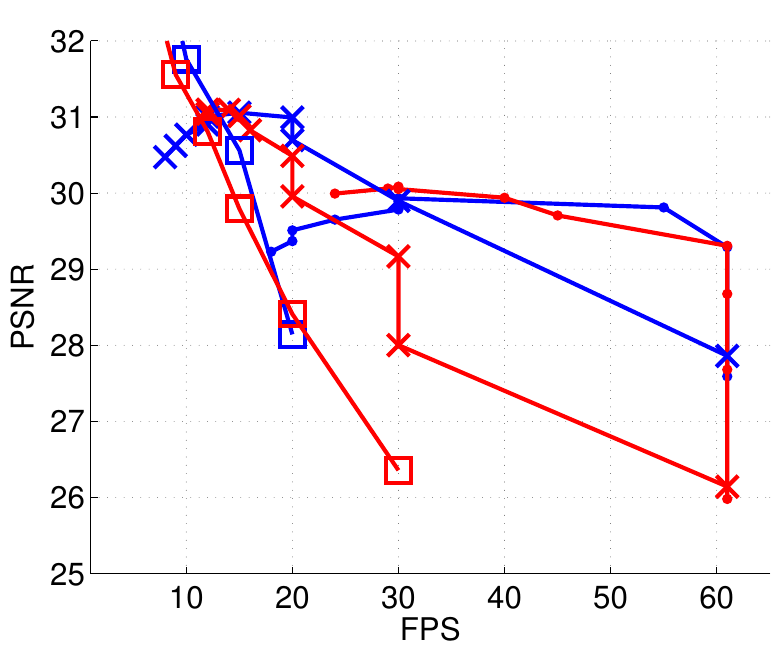}%
    \llap{\raisebox{3.5cm}{
      \includegraphics[width=0.22\textwidth]{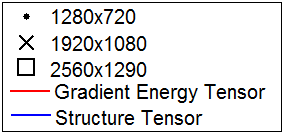}%
    }}
	\hfill  \\
	\hfill (c) \hfill \hfill (d) \hfill \phantom{I} \\
	\caption{Obtained PSNR (a), SSIM (b) and fps (c) for the considered image sizes. The PSNR and SSIM values are similar for the two tensors but the obtained fps is significantly higher for the gradient energy tensor.}
	\label{fig:iteration_fps}
\end{figure*}



Figure~\ref{fig:iteration_fps} show the obtained PSNR (a) and SSIM (b) values compared to the iteration number, as expected the two methods are comparable for the obtained error measures. The best error values are in agreement between the two methods but obtained at different iterations numbers, it is not a discrepancy in method performance but an issue of parameter tuning. In figure~\ref{fig:iteration_fps} (c) and (d) we show the fps that we obtain for each iteration. After four iterations, in (c), the filter using the gradient energy tensor is stable at 60 fps whereas the standard diffusion scheme using the structure tensor has dropped to 30 fps for the smallest image resolution. In (d) we show the tradeoff between PSNR and obtained fps for \emph{both} tensors: a higher fps result in a lower PSNR value. Note that the fps count is \emph{independent} of the image content. Future work will include a more comprehensive study of the GET orientation estimation compared to the structure tensor for other noise levels and image types than considered in this work. 

\begin{figure*}[t]
   \centering
   \hfill
   \includegraphics[width=0.24\textwidth]{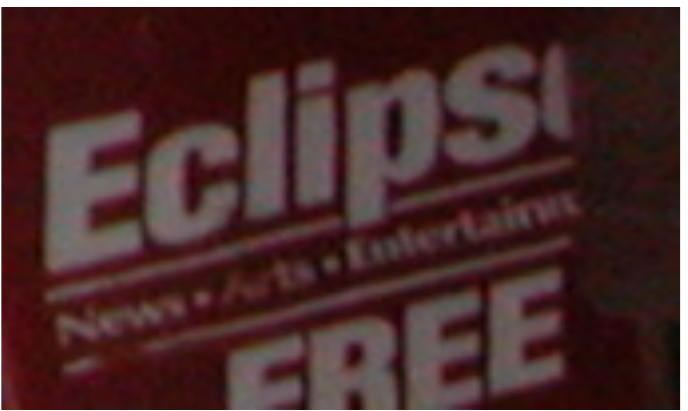} 
   \hfill\hfill
   \includegraphics[width=0.24\textwidth]{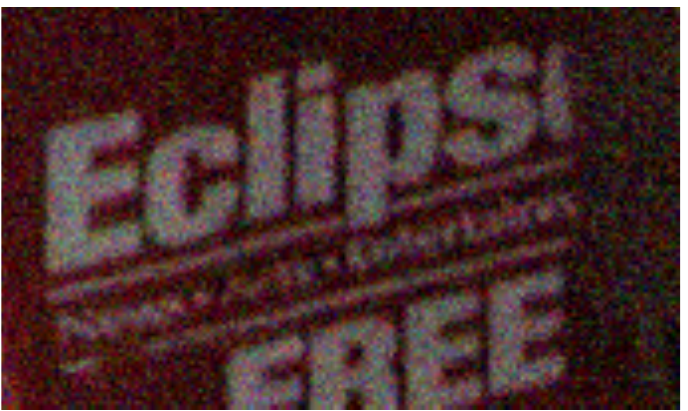} 
   \hfill\hfill
   \includegraphics[width=0.24\textwidth]{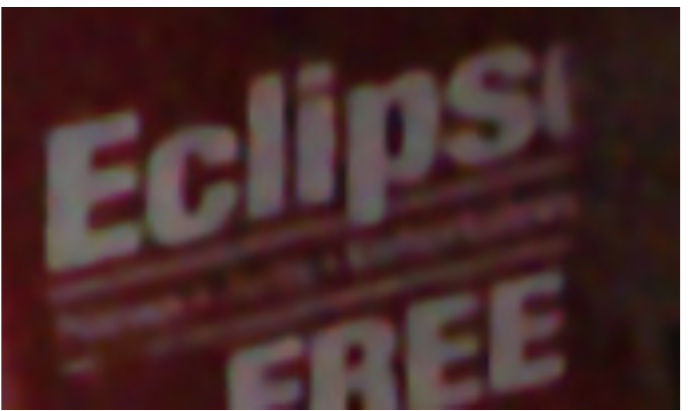} 
   \hfill\hfill
   \includegraphics[width=0.24\textwidth]{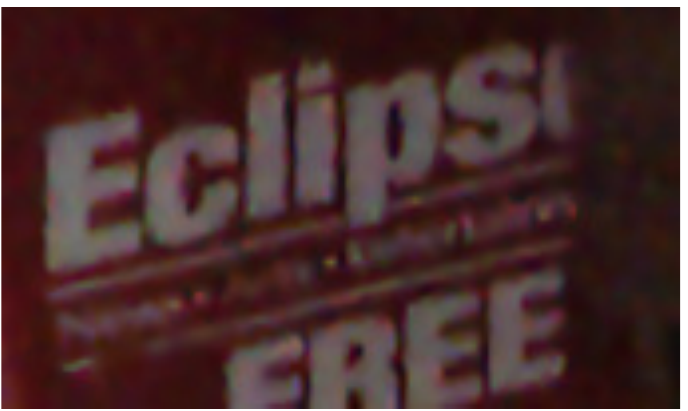} \hfill \\
   \hfill
   \includegraphics[width=0.24\textwidth]{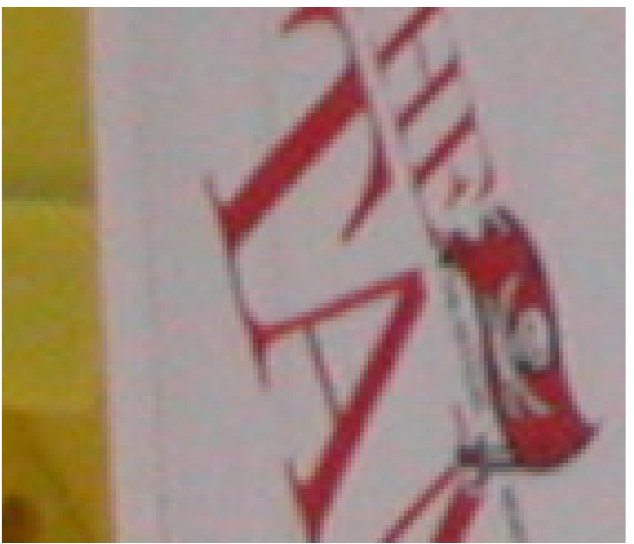} 
   \hfill\hfill
   \includegraphics[width=0.24\textwidth]{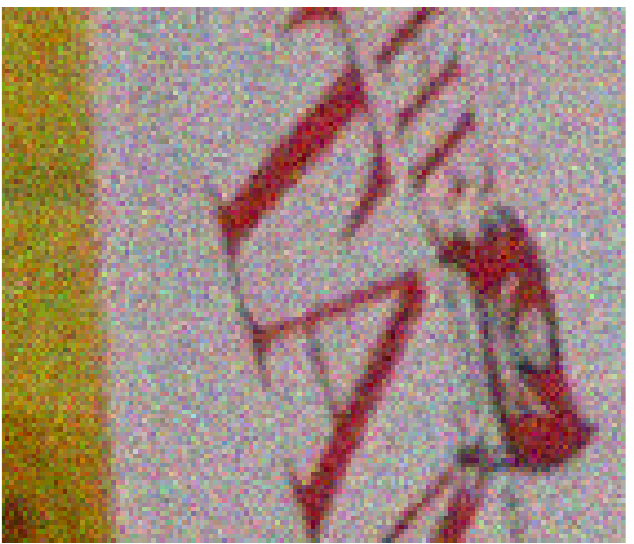} 
   \hfill\hfill
   \includegraphics[width=0.24\textwidth]{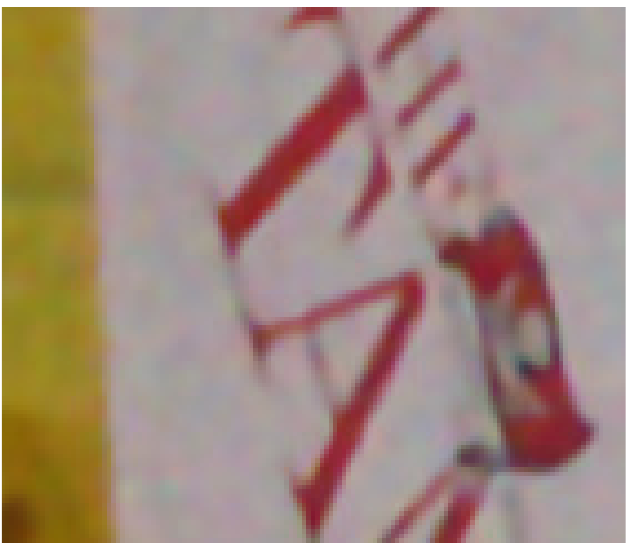} 
   \hfill\hfill
   \includegraphics[width=0.24\textwidth]{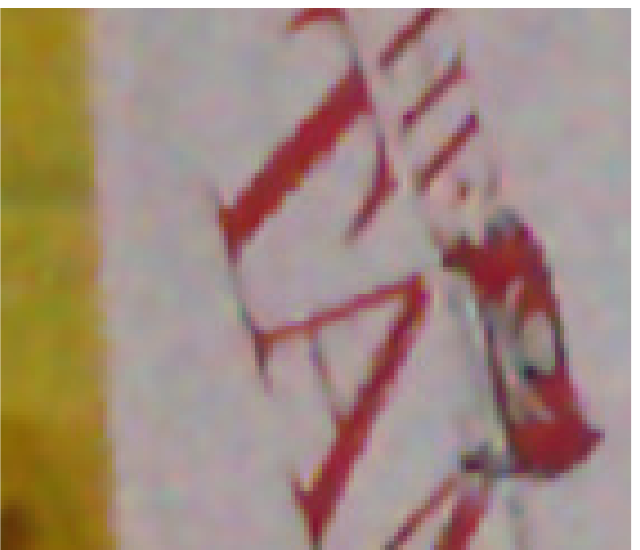} \hfill \\
	\begin{tabular}{m{3cm} m{3cm} m{3cm} m{3cm}}
	Original & Noisy & Structure tensor & Energy Tensor \\
	\end{tabular}
   \caption{Two patches from the $2560\times 1920$ resolution image and the corresponding denoised images at iteration 10. The applied noise was 20 standard deviation of normal distributed additive noise. Note that the difference in visual appearance is not discernible but the fps count is improved by 25\% with the gradient energy tensor.}
	\label{fig:test_images_result}
\end{figure*}

In figure~\ref{fig:test_images_result} we illustrate an example of the final denoised result, the zoomed images were cropped from the $2560\times 1920$ resolution image and the original image was corrupted by 20 standard deviations of normal distributed additive noise. The images illustrate the denoised result after 10 iterations and note that the gradient energy tensor fps-count is 25\% higher than the structure tensor but with indiscernible image quality and PSNR (cmp. figure~\ref{fig:iteration_fps} (a) and (c)). In table~\ref{table:ratios} we give the ratio of the gradient energy tensor error measures over the structure tensor error measures averaged over the iterations, \ie if the ratio is identical to 1.0 there is no difference in method performance, if the value is larger than 1.0 then the ratio indicate that the gradient energy tensor performs better. The SSIM value difference is less than $10^{-3}$  whereas the PSNR shows a 3.2\% difference in the favour of the structure tensor, however considering the final fps-count the gradient energy tensor is up to 40\% more computationally efficient for the largest image size. 

\begin{table}[t]
	\centering
	\caption{Ratios are computed as the measure obtained by the gradient energy tensor divided by the measure obtained by the structure tensor and averaged over iterations. A ratio of 1.0 show that the performance is identical. The SSIM and PSNR ratios are nearly identical but at up to 40\% higher fps values in the largest image resolution, the PSNR show a marginal loss in image quality for the gradient energy tensor.}
	\begin{tabular*}{0.5\textwidth}{@{\extracolsep{\fill} }  c | ccc }
		\emph{Ratio}		& PSNR 	& SSIM 	& fps 	\\
		\hline
		$1280\times 720$   	& 0.991	& 0.995	& 1.35	\\
		$1920\times 1080$   & 0.982 & 0.997	& 1.27	\\
		$2560\times 1920$   & 0.968	& 0.999	& 1.40	\\
	\end{tabular*}
	\label{table:ratios}
\end{table}

\section{Conclusion}
In this work we have presented the gradient energy tensor as an alternative to the structure tensor for local orientation. We have considered three applications: corner detection, optical flow and adaptive image enhancement. Due to the absence of a post-convolution of the gradient energy tensor components the tensor is highly suitable for efficient implementation on the GPU, and we showed that the tensor yield significant improvement in obtained frames per second compared to the structure tensor without compromising PSNR and SSIM error values.

\vspace{3mm}
\noindent {\bf Acknowledgement}. This research has received funding from the Swedish Research Council through grants for the projects Visualization-adaptive Iterative Denoising of Images (VIDI) and Extended Target Tracking (ETT), within the Linnaeus environment CADICS and the excellence network ELLIIT.


\bibliographystyle{splncs}
\bibliography{ACCV2014}

\end{document}